\documentclass[10pt]{article} % For LaTeX2e

\usepackage[preprint]{rlj} % Should be uncommented for preprint versions

\usepackage{amssymb}            % Defines common symbols like \mathbb R
\usepackage{mathtools}          % Extends amsmath, providing common math tools
\usepackage{mathrsfs}           % Enables \mathscr, which can work in cases that \mathcal does not
\usepackage{graphicx}           % For including images
\usepackage{subcaption}         % Allows for the use of subfigures and subcaptions
\usepackage[space]{grffile}     % For spaces in image names
\usepackage{url}                % For displaying URLs
\usepackage{lipsum}             % For placeholder text
\usepackage{booktabs}
\usepackage{algorithm}
\usepackage{algorithmic}

\newcommand{\nomealgo}{SOPE}
\title{SOPE: Stabilizing Off-Policy Evaluation for Online RL with Prior Data}

\setrunningtitle{Stabilizing Off-Policy Evaluation for Online RL with prior data}

\author{Carlo Romeo\textsuperscript{$\dagger$ 1}, Girolamo Macaluso\textsuperscript{$\dagger$ 2}, Alessandro Sestini\textsuperscript{3}, Andrew D. Bagdanov\textsuperscript{2}}

\coverPageAuthor{Carlo Romeo\textsuperscript$\dagger$, Girolamo Macaluso\textsuperscript$\dagger$, Alessandro Sestini, Andrew D. Bagdanov}

\emails{
\textsuperscript{1}\{carlo.romeo, girolamo.macaluso, andrew.bagdanov\}@unifi.it \\
\textsuperscript{2}asestini@ea.com}

\affiliations{
$^{1}$\textbf{Media Integration and Communication Center – University of Florence}\\
$^{2}$\textbf{SEED – Electronic Arts}\\
\par % If including additional comments like below, use \par to add some whitespace. 
$^\dagger$ The authors contributed equally as co-first authors.
}

\contribution{
    We introduce \nomealgo, a novel algorithm for online Reinforcement Learning with prior data that employs Off-Policy Policy Evaluation as an automated early-stopping mechanism to dynamically control the length of offline training phases without manual tuning.
    }
    {
    Prior stabilization-based methods rely on rigid, task-dependent hyperparameters for offline update schedules, which require extensive and brittle manual tuning.
    }

\contribution{
    We demonstrate that \nomealgo~achieves cutting-edge performance, delivering top performance with up to 45.6\% improvement in 25 different continuous control tasks from the Minari benchmark suite.
    }
    {
    The comprehensive evaluation encompasses expert, medium, and simple dataset qualities across MuJoCo environments, outperforming existing baselines like RLPD, Cal-QL, SPEQ, and SACfD.
    }

\contribution{
    We show that \nomealgo~yields massive reductions in computational cost, achieving up to $36.8\times$ reduction in TFLOPs compared to leading compute-heavy baselines.
    }
    {
    High Update-to-Data algorithms and multi-stage offline-to-online methods typically incur immense computational overhead to bridge the distributional gap and achieve sample efficiency.
    }

 % Your keywords

\summary{
Incorporating prior data into online reinforcement learning accelerates training but typically forces a difficult trade-off between high computational costs and long, multi-stage training pipelines. While fixed-length stabilization phases are significantly more computationally efficient than static update schedules, they require task-dependent manual tuning, risking either the waste of prior knowledge or severe overfitting. To address this, we propose \nomealgo, an algorithm that uses an actor-aligned Off-Policy Policy Evaluation (OPE) signal as an automated early-stopping mechanism to dynamically control the length of offline training phases. By evaluating the critic on a held-out validation split under the current policy's action distribution, \nomealgo~halts gradient updates exactly when out-of-distribution benefits saturate, eliminating the need for manual schedule tuning. 
Evaluated on 25 continuous control tasks from the Minari benchmark suite, our results demonstrate that adaptive, evaluation-driven update schedules is more effective than relying on static, exhaustive update schedules.
}

\begin{document}

% \makeCover  % Create the cover page
\maketitle  % Make the title section

\begin{abstract}
Incorporating prior data into online reinforcement learning accelerates training but typically forces a difficult trade-off between high computational costs and long, multi-stage training pipelines. While fixed-length stabilization phases are significantly more computationally efficient than static update schedules, they require task-dependent manual tuning, risking either the waste of prior knowledge or severe overfitting. To address this, we propose \nomealgo, an algorithm that uses an actor-aligned Off-Policy Policy Evaluation (OPE) signal as an automated early-stopping mechanism to dynamically control the length of offline training phases. By evaluating the critic on a held-out validation split under the current policy's action distribution, \nomealgo~halts gradient updates exactly when out-of-distribution benefits saturate, eliminating the need for manual schedule tuning. 
Evaluated on 25 continuous control tasks from the Minari benchmark suite, \nomealgo~improves baseline performance by up to 45.6\% while reducing the required TFLOPs by up to 22$\times$, thus balancing the tradeoff between sample and computational efficiency. These findings demonstrate that adaptive, evaluation-driven update schedules is more effective than relying on static, exhaustive update schedules.
Source code is available at \href{https://github.com/CarloRomeo427/SOPE.git}{github.com/CarloRomeo427/SOPE}.

\end{abstract}

\section{Introduction}
Reinforcement Learning~\citep{sutton, survey} (RL) has achieved impressive results across diverse domains \citep{berner2019dota, mathieu2023alphastar, bergdahl2024reinforcement, black2023training, guo2025deepseek}. However, its demanding sample complexity often limits real-world applicability. To accelerate learning, agents can leverage pre-collected datasets (e.g., human demonstrations or logged interactions) during training, a paradigm known as \emph{online reinforcement learning with prior data}~\citep{ball2023efficient, song2022hybrid, nair2020awac}. Unlike strict offline-to-online approaches that require an expensive and distinct pre-training phase~\citep{ shin2025online, lei2023uni, nakamoto2023cal}, RL with prior data treats prior data as a continuous resource available during online learning.

A core challenge in this setting is leveraging the offline data in an efficient way that also mitigates the problems of distributional shift and value overestimation~\citep{levine2020offline, kumar2020conservative}. The solution proposed by RLPD~\citep{ball2023efficient} combines: high \emph{Update-to-Data} (UTD) ratio~\citep{chen2021randomized, hiraoka2021dropout} for high sample efficiency; symmetric sampling to include the offline data directly into the online training process while minimizing distributional shift; and a large ensemble of critic functions to reduce overestimation. A major drawback of this approach is its computational cost. Due to its use of a large critic ensemble and its high UTD, it requires significantly long training times and computational resources. A growing body of work in online RL aims to alleviate such computational burdens. 
% Approaches like SPEQ~\citep{romeo2025speq} alternate online learning phases with intensive \textit{offline stabilization} phases which reduce overall training cost while improving performance. 
% However, the benefit gained from offline stabilization phases is counterbalanced by the introduction of a task-dependent hyperparameter: the length of stabilization phases. A fixed length is often suboptimal because under-stabilization wastes prior knowledge, while over-stabilization leads to overfitting.
SPEQ \citep{romeo2025speq} alternates online learning with offline stabilization to reduce costs and improve performance. However, it introduces a sensitive, task-dependent hyperparameter: stabilization length.

In this paper, we introduce \textbf{S}tabilizing \textbf{O}ff-\textbf{P}olicy \textbf{E}valuation for Online RL with prior data (\nomealgo), a novel algorithm that uses Off-Policy Policy Evaluation (OPE)~\citep{dudik2011doubly, voloshin2021empirical} as an automated early-stopping mechanism during offline stabilization phases. 
Our experiments show that, by using OPE as a termination signal, SOPE is able to dynamically control the length of each offline stabilization phase given the current data distribution, without requiring manual per-task tuning. SOPE replaces the stabilization length hyperparameter with a more robust patience hyperparameter which controls early stopping during offline stabilization. Our empirical evaluation across 25 diverse online RL with prior data Mujoco continuous control tasks~\citep{mujoco} from the Minari benchmark suite~\citep{minari} demonstrates that \nomealgo~achieves state-of-the-art performance ($77.94\%$ aggregate normalized score). Crucially, it does so with a highly efficient computational footprint, yielding a $36.8\times$ and $8.2\times$ reduction in TFLOPs compared to leading baselines like RLPD and Cal-QL, respectively. 

In summary, our key contributions are: 
\begin{itemize} 
    \item We introduce \nomealgo, an algorithm that leverages OPE to dynamically halt gradient updates during offline stabilization phases. 
    \item We demonstrate that \nomealgo~ improves performance by 45.6\% and 13.8\%, respectively, over state-of-the-art benchmarks such as RLPD and Cal-QL. This superior performance has been validated on 25 different continuous control tasks from the Minari benchmark suite.
    \item We show that \nomealgo~achieves high sample efficiency while consuming only a fraction of the compute required by existing methods. Specifically, it yields a $36.8\times$ and $8.2\times$ reduction in TFLOPs compared to baselines like RLPD and Cal-QL, respectively.
\end{itemize}

\section{Related Work}  
\label{sec:related}
In this section we review the current state-of-the-art approaches for online RL with prior data and discuss recent advances in computationally efficient and sample-efficient online RL.

\textbf{Online Reinforcement Learning with Prior Data.}
In many practical applications, agents leverage pre-collected datasets to accelerate online learning. Existing approaches generally fall into two categories. The first consists of \emph{offline-to-online} methods, such as Cal-QL~\citep{nakamoto2023cal}, Uni-O4~\citep{lei2023uni}, and OPT~\citep{shin2025online}. These rely on multi-stage protocols -- typically extensive offline pre-training followed by online fine-tuning -- to bridge the distributional gap between data sources.

The second category, known as \emph{online RL with Prior Data}~\citep{ball2023efficient} or \emph{hybrid RL}~\citep{song2022hybrid}, avoids the pre-training bottleneck by treating the offline dataset as a resource during online training. While naive extensions like SACfD~\citep{sacfd1, nair2020awac} often result in offline data overwhelming new online experiences, methods like RLPD~\citep{ball2023efficient} and A3RL~\citep{a3rl} counter this using symmetric sampling, high UTD ratios, and large ensembles. However, this significantly increases computational costs. 
\nomealgo~is a single-stage approach that integrates prior knowledge to achieve high sample efficiency while minimizing computational overhead by replacing the high-UTD training process with offline stabilization phases.

\textbf{Sample-Efficient Online Reinforcement Learning.}
A primary strategy for improving sample efficiency is increasing the Update-To-Data (UTD) ratio. While model-based methods like MBPO~\citep{mbpo} and model-free approaches like REDQ~\citep{chen2021randomized}, DroQ~\citep{droq}, and SMR~\citep{smr} successfully accelerate learning via high UTD ratios, their reliance on a continuous volume of updates requires substantial computational resources. SPEQ~\citep{romeo2025speq} mitigates this computational burden by, instead of using a constant UTD ratio, alternating between online learning phases and offline stabilization phases. 
% During stabilization phases the policy network is frozen while critic updates are performed using the static replay buffer. 
However, SPEQ relies on a task-specific hyperparameter that defines the length of these stabilization phases. \nomealgo~avoids manual tuning of this hyperparameter by dynamically determining the necessary number of updates during each offline phase.

\section{Automated Stabilization Phases via Off-policy Policy Evaluation}
In this section we describe problem setting for online RL with prior data, analyze the computational limitations of the current state-of-the-art, motivate the need for an automated training schedule, and introduce \nomealgo, which leverages OPE to adaptively determine the optimal length of offline stabilization phases through an actor-aligned stopping criterion.

\textbf{Problem Setting.}
We consider a Markov Decision Process~\citep{bellman1957markovian} in which the agent seeks a policy $\pi$ maximizing:
\begin{equation}
    J(\pi) = \mathbb{E}_{\pi}\!\left[\sum_{t=0}^{\infty} \gamma^t r(s_t, a_t)\right]. 
\end{equation}
In the \emph{online RL with prior data} setting~\citep{ball2023efficient, song2022hybrid, sacfd1, nair2020awac}, the agent has access to a pre-collected offline dataset $\mathcal{D}_{\text{off}} = \{(s, a, r, s')\}$ generated by unknown behavior policies. The limited and unknown nature of the behavior policy often results in function approximation overestimating Q-values for out-of-distribution inputs~\citep{fujimoto2019off}.
Unlike the purely offline regime, however, the agent can additionally collect online transitions $\mathcal{D}_{\text{on}}$, progressively correcting distributional shift. The central challenge is to leverage $\mathcal{D}_{\text{off}}$ to accelerate online learning while mitigating value overestimation.

% ──────────────────────────────────────────────────────────────────────────────
\textbf{Limitation of current approaches.}
RLPD~\citep{ball2023efficient} extends SAC~\citep{haarnoja2018soft} to incorporate prior data through three mechanisms:
\emph{(i) symmetric sampling}, in which each mini-batch $\mathcal{B} = \mathcal{B}_{\text{off}} \cup \mathcal{B}_{\text{on}}$ with $|\mathcal{B}_{\text{off}}| = |\mathcal{B}_{\text{on}}| = |\mathcal{B}|/2$ draws half of the transitions from $\mathcal{D}_{\text{off}}$ and half from $\mathcal{D}_{\text{on}}$;
\emph{(ii) ensemble critics}, maintaining $E{=}10$ Q-functions $\{Q_{\phi_i}\}_{i=1}^{E}$ with Layer Normalization~\citep{ba2016layer} to bound extrapolation, where TD targets use a random subset $\mathcal{Z} \subset \{1,\dots,E\}$ as $y = r + \gamma \min_{i \in \mathcal{Z}} Q_{\phi'_i}(s', \tilde{a}')$ with $\tilde{a}' \sim \pi_\theta(\cdot \mid s')$;
and \emph{(iii) a high and fixed UTD ratio} performing more gradient updates per environment step on critics ($UTD = 20)$.
As already highlighted in Section~\ref{sec:related}, this combination requires substantial computational resources: the total number of critic gradient updates is up to two orders of magnitude more expensive than standard SAC~\citep{haarnoja2018soft}.

SPEQ~\citep{romeo2025speq} addresses this bottleneck by restructuring the update schedule. High-UTD algorithms perform many updates between every consecutive environment interaction, repeatedly updating on a replay buffer that has grown by only a single transition. SPEQ instead decouples training into two alternating phases, allowing the buffer to grow before investing in critic refinement.
During training, SPEQ alternates between \emph{online phases}, which operates as standard SAC, and \emph{offline stabilization phases} in which online interaction is paused, the actor $\pi_\theta$ is frozen, and the critics $Q_{\phi_1}$ and $Q_{\phi_2}$ are updated for $N$ consecutive gradient steps on the fixed replay buffer. Dropout regularization~\citep{dropout} mitigates overestimation, providing implicit ensembling without large critic sets~\citep{chen2021randomized, droq}.

%% ──────────────────────────────────────────────────────────────────────────────
\textbf{Problems with fixed-length stabilization.}
\label{sec:brittleness}
While SPEQ significantly reduces computational cost compared to high-UTD approaches, it introduces a new hyperparameter: the stabilization length $N$. Achieving optimal performance requires extensive cross-validation, as $N$ is highly environment dependent.
\begin{figure}[t]
    \centering
        \includegraphics[width=0.5\textwidth]{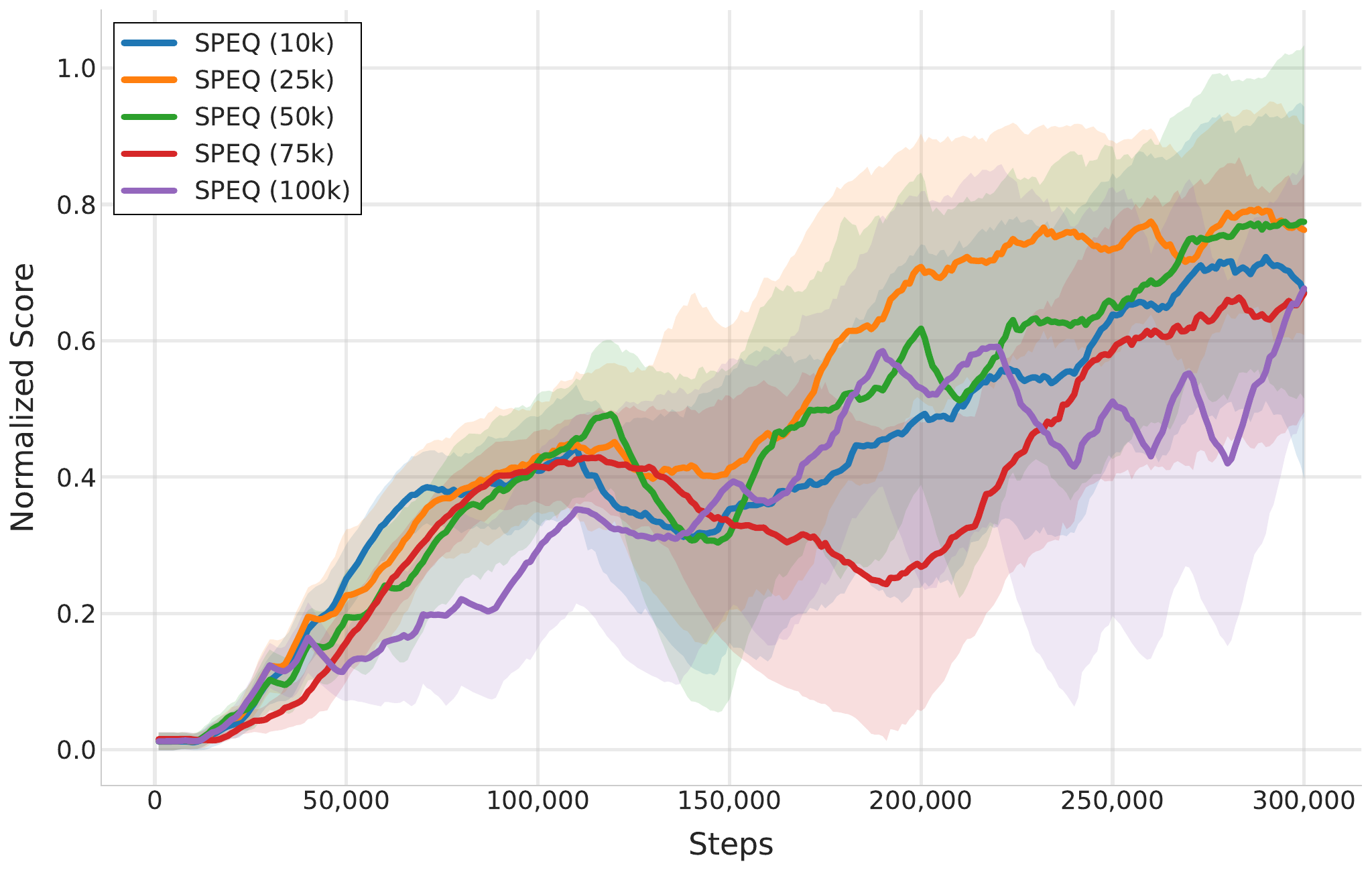}
\caption{\textbf{Sensitivity of SPEQ to stabilization length on HalfCheetah.} Final performance does not scale monotonically with the number of updates. Increasing the budget from 10k (blue) to the default 75k setting (red) results in similar performance despite a $7\times$ increase in computation, while intermediate values like 25k (orange) and 50k (green) achieve superior results.
}
\label{fig:speq_brittle}
\end{figure}
Figure~\ref{fig:speq_brittle} shows multiple SPEQ training runs on the HalfCheetah task, varying the number $N$ of updates per stabilization phase. The plot highlights the sensitivity of SPEQ to this hyperparameter. The default configuration (red curve) is significantly outperformed by less computationally intensive settings, such as 25k (orange) and 50k (green) updates. Most notably, the 10k configuration (blue) matches the default performance while requiring 7$\times$ less computation. This result suggests that a static $N$ often leads to diminishing returns or even performance degradation. To address the task-dependent nature of $N$ and mitigate this computational burden, we propose an adaptive mechanism to adaptively select the stabilization phase length.

% ──────────────────────────────────────────────────────────────────────────────
\textbf{Stabilization phases as Off-policy Policy Evaluation.}
\label{sec:ope}
Results from the OPE literature show a characteristic bias-variance trade-off: while initial fitting iterations reduce approximation bias, prolonged optimization on finite data eventually increases estimation variance as the model overfits to training samples~\citep{munos2008finite, kumar2020conservative, levine2020offline}. This implies that the marginal utility of additional critic updates is diminishing and, beyond a certain threshold, potentially negative for policy improvement. Stabilization-length selection is therefore naturally an early-stopping problem: we seek the point along the critic's optimization trajectory after which further updates stop to improve and may begin to harm out-of-distribution performance.

\textbf{Actor-aligned evaluation.}
A crucial advantage of the online RL with prior data paradigm is the availability of pre-collected transitions. This data allows us to hold out a portion of the replay buffer strictly for evaluation.
At the beginning of each stabilization phase, we partition the current replay buffer $\mathcal{D} = \mathcal{D}_{\text{off}} \cup \mathcal{D}_{\text{on}}$ into two disjoint splits, $\mathcal{D}_{\mathrm{train}}$ and $\mathcal{D}_{\mathrm{val}}$
We update the critics only on $\mathcal{D}_{\text{train}}$ and reserve $\mathcal{D}_{\text{val}}$ to decide when to stop.

We define our stopping criterion as ``\textit{actor-aligned}'' as it measures critic quality under the policy's own action distribution (i.e. the distribution that will be queried during subsequent policy improvements). To define our stopping criterion, we use a Direct Method (DM) estimator~\citep{dudik2011doubly, voloshin2021empirical}. For the held-out validation set $\mathcal{D}_{\mathrm{val}}$, we define:
\begin{equation}
\widehat{J}_{\mathrm{DM}}(\pi_\theta;\phi) \;:=\; \mathbb{E}_{s\sim \mathcal{D}_{\mathrm{val}}} \mathbb{E}_{a\sim \pi_\theta(\cdot|s)} \left[Q_{\phi}(s,a)\right].
\label{eq:j_dm}
\end{equation}
Intuitively, $\widehat{J}_{\mathrm{DM}}$ increases when critic predictions for on-policy actions improve in a way that generalizes beyond $\mathcal{D}_{\text{train}}$, and it plateaus or decreases when further critic fitting overfits the training split~\citep{kumar2020conservative, paine2020hyperparameter}. 
However, $\widehat{J}_{\mathrm{DM}}$ is not intended as an unbiased estimator of the true environment returns. Rather, it serves as a proxy for whether the stabilization phase is still improving the critic functions on out-of-distribution samples under $\pi_\theta$~\citep{farajtabar2018more}.

DM estimators can exhibit significant bias when used as absolute return estimators~\citep{voloshin2021empirical}. However, in our case, we use $\widehat{J}_{\mathrm{DM}}$ only to compare critics along a single optimization trajectory within a specific stabilization phase. For such relative comparisons, systematic biases that remain approximately constant across successive critic iterations cancel and do not affect the selection of the optimal stabilization length. This observation aligns with recent work demonstrating that biased OPE estimators for model selection, provided that the bias is approximately shared across the candidates being compared~\citep{paine2020hyperparameter, zhang2021towards}.

% ──────────────────────────────────────────────────────────────────────────────
\textbf{The \nomealgo~Algorithm.}
\label{sec:algo}
The core of \nomealgo~is an adaptive schedule that modulates the computational effort of offline stabilization phases. The agent consists of an actor $\pi_\theta$ and a pair of critics $Q_{\phi_1},Q_{\phi_2}$.
As in SPEQ~\citep{romeo2025speq}, training alternates between a standard online phase in which $\pi_\theta$ is trained using SAC~\citep{haarnoja2018soft} and an offline stabilization phase. During a stabilization phase, we freeze the actor parameters $\theta$ and update only the critic parameters $\phi_{1,2}$ on a fixed dataset drawn from $\mathcal{D}$ using symmetric sampling~\citep{ball2023efficient}. We monitor the actor-aligned proxy using $\widehat{J}_{\mathrm{DM}}$ and stop training when improvement plateaus. To ensure this mechanism is robust to the high-frequency stochasticity of RL learning, we introduce a patience-based stopping rule: we define patience $P$ as the number of consecutive update steps used to determine whether performance recovers following an initial drop. This patience window prevents the stabilization from terminating prematurely due to random noise, while still reliably halting computation once the out-of-distribution gains have truly saturated. We show in Section~\ref{sec:experiments} that this early-stopping rule is robust to the patience value chosen and requires no fine-grained, manual tuning. Further details and pseudo-code are provided in Section~\ref{supp:algo} of the Supplementary Materials.

\section{Experimental Results\protect\footnote{Code to reproduce our results will be released upon publication.}}
\label{sec:experiments}

We evaluate \nomealgo~using the Minari~\citep{minari} benchmark suite, which provides standardized offline datasets for Gymnasium-MuJoCo~\citep{gymnasium, mujoco} environments. Each dataset has three quality levels: \textit{expert}, \textit{medium}, and \textit{simple}. The \textit{medium} and \textit{expert} datasets are available for all 10 supported environments (Humanoid, Ant, HalfCheetah, Hopper, Walker2d, InvertedPendulum, InvertedDoublePendulum, Pusher, Reacher, and Swimmer), whereas the \textit{simple} datasets are available only for the 5 standard locomotion tasks. This results in 25 unique environment-dataset combinations. The \textit{expert} datasets consist of trajectories collected using a policy trained to full convergence; \textit{medium} datasets are generated by collecting transitions from a partially trained policy; and \textit{simple} datasets are collected from a policy in the early stages of training. 

We benchmark \nomealgo~against: online RL with prior data algorithms such as \textbf{RLPD}~\citep{ball2023efficient}, \textbf{SACfD}~\citep{sacfd1}; an offline to online approach \textbf{Cal-QL}~\citep{nakamoto2023cal}; and \textbf{SPEQ O2O}~\citep{romeo2025speq}, which is an adaptation of the original SPEQ algorithm for online RL with prior data. We modify it through the use of symmetric sampling: while keeping the original SPEQ's structure, we use an external offline dataset and we sample 50\% of the batch from the online dataset and 50\% from the offline one, in both the online and offline stabilization phases. Although A3RL~\citep{a3rl} extends the RLPD framework, it is omitted from this evaluation due to the absence of an official open-source implementation. 
All algorithms are trained for 300,000 online environment steps. Cal-QL is the only algorithm that requires an additional offline pretraining step consisting of 1 million updates using the offline dataset. We report the average normalized score (where 0 represents random performance and 100 represents expert performance) across 10 random seeds for each environment-dataset combination.
To facilitate future research, we have released the SPEQ code-base here: \href{https://github.com/CarloRomeo427/SOPE.git}{github.com/CarloRomeo427/SOPE}.

Our experimental analysis aims to address the following research questions:
\begin{itemize}
    \item \textbf{Q1:} Is dynamic allocation of offline updates superior to fixed schedules?
    \item \textbf{Q2:} Is \nomealgo~robust to the choice of the patience hyperparameter $P$?
    \item \textbf{Q3:} How does the algorithm distribute offline updates across the training process?
    \item \textbf{Q4:} What is the computational efficiency of SOPE compared to baselines?
\end{itemize}

\textbf{Q1: Is dynamic allocation of offline updates superior to fixed schedules?}
We evaluate all baselines across the full experimental suite. Table~\ref{tab:results_all} provides a summary of the normalized scores for each method. Referring to the aggregate normalized scores, \nomealgo~achieves the highest overall performance ($\approx 77\%$), outperforming SPEQ O2O ($\approx 72\%$), Cal-QL ($\approx 71\%$), SACfD ($\approx 67\%$), and RLPD ($\approx 53\%$). A notable finding is the underperformance of RLPD. This suggests that the computational overhead and high-UTD strategy of RLPD do not generalize robustly across heterogeneous task distributions, confirming the hypothesis that uniform high-frequency updating may yield diminishing returns or instability in diverse settings. See the Supplementary Materials (Section~\ref{supp:experiments}) for extended the results, including aggregated (Figure~\ref{supp:aggregated}) and individual (Figure~\ref{supp:exploded_views}) plots.

\begin{table*}[t]
\centering
\caption{Normalized scores (\%) on \textbf{expert}, \textbf{medium}, and \textbf{simple} datasets (mean $\pm$ std across 10 seeds). Best per environment in bold. Values reported with $\pm 0.0$ denote a standard deviation of less than $0.1\%$, representing minimal but non-zero fluctuation across seeds.}
\label{tab:results_all}
\resizebox{0.95\textwidth}{!}{
\begin{tabular}{lccccc}
\toprule
Environment & SPEQ O2O & RLPD & Cal-QL & SACfD & \textbf{\nomealgo~(OURS)} \\
\midrule
\multicolumn{6}{c}{\textbf{Expert Dataset}} \\
\midrule
Humanoid & $\phantom{00}4.7 \pm \phantom{0}1.2$ & $\mathbf{\phantom{0}50.0 \pm 23.2}$ & $\phantom{00}6.2 \pm \phantom{0}3.7$ & $\phantom{0}19.1 \pm 13.8$ & $\phantom{00}4.7 \pm \phantom{0}0.8$ \\
Ant & $\phantom{0}95.4 \pm \phantom{0}4.3$ & $\phantom{00}1.0 \pm \phantom{0}0.1$ & $\phantom{0}93.0 \pm \phantom{0}4.3$ & $\phantom{0}72.9 \pm 24.9$ & $\mathbf{100.1 \pm \phantom{0}3.2}$ \\
HalfCheetah & $\phantom{0}72.1 \pm 15.4$ & $\phantom{0}34.8 \pm \phantom{0}9.5$ & $\phantom{0}50.1 \pm 24.2$ & $\phantom{0}43.4 \pm 43.1$ & $\mathbf{\phantom{0}76.8 \pm 15.1}$ \\
Hopper & $\phantom{0}41.6 \pm 13.0$ & $\mathbf{\phantom{0}63.8 \pm 28.8}$ & $\phantom{0}51.3 \pm 15.6$ & $\phantom{0}42.0 \pm 16.3$ & $\phantom{0}50.5 \pm 19.1$ \\
Walker2d & $\phantom{0}55.5 \pm 22.6$ & $\phantom{0}39.8 \pm 18.2$ & $\mathbf{\phantom{0}73.5 \pm 11.9}$ & $\phantom{0}43.5 \pm 36.3$ & $\phantom{0}66.1 \pm 21.2$ \\
InvertedPendulum & $\mathbf{100.0 \pm \phantom{0}0.0}$ & $100.0 \pm \phantom{0}0.0$ & $\phantom{0}98.2 \pm \phantom{0}3.1$ & $\phantom{0}98.7 \pm \phantom{0}3.1$ & $100.0 \pm \phantom{0}0.0$ \\
InvertedDoublePendulum & $\mathbf{100.0 \pm \phantom{0}0.1}$ & $\phantom{0}99.9 \pm \phantom{0}0.5$ & $\phantom{0}91.2 \pm \phantom{0}7.5$ & $\phantom{0}99.3 \pm \phantom{0}1.7$ & $\phantom{0}99.3 \pm \phantom{0}1.2$ \\
Pusher & $\phantom{0}99.6 \pm \phantom{0}1.1$ & $\phantom{0}96.2 \pm \phantom{0}2.1$ & $\phantom{0}99.3 \pm \phantom{0}1.2$ & $\phantom{0}99.1 \pm \phantom{0}1.2$ & $\mathbf{\phantom{0}99.9 \pm \phantom{0}1.0}$ \\
Reacher & $102.7 \pm \phantom{0}0.0$ & $101.9 \pm \phantom{0}1.3$ & $\mathbf{102.8 \pm \phantom{0}1.0}$ & $102.6 \pm \phantom{0}1.0$ & $102.2 \pm \phantom{0}1.1$ \\
Swimmer & $\phantom{0}97.5 \pm \phantom{0}0.0$ & $\phantom{0}13.4 \pm \phantom{0}1.3$ & $\phantom{0}12.7 \pm \phantom{0}0.3$ & $\phantom{0}97.2 \pm \phantom{0}2.3$ & $\mathbf{\phantom{0}98.1 \pm \phantom{0}0.4}$ \\
\textbf{Expert Average} & $\phantom{0}76.9 \pm 33.3$ & $\phantom{0}60.1 \pm 38.1$ & $\phantom{0}67.8 \pm 36.2$ & $\phantom{0}71.8 \pm 31.8$ & $\mathbf{\phantom{0}79.8 \pm 31.9}$ \\
\midrule
\multicolumn{6}{c}{\textbf{Medium Dataset}} \\
\midrule
Humanoid & $\phantom{0}50.9 \pm 20.3$ & $\phantom{0}44.7 \pm 25.1$ & $\mathbf{\phantom{0}78.3 \pm \phantom{0}8.0}$ & $\phantom{0}14.9 \pm \phantom{0}4.4$ & $\phantom{0}66.2 \pm 10.7$ \\
Ant & $\phantom{0}86.8 \pm \phantom{0}7.7$ & $\phantom{00}1.0 \pm \phantom{0}0.0$ & $\phantom{0}86.0 \pm \phantom{0}6.9$ & $\phantom{0}76.3 \pm 14.3$ & $\mathbf{\phantom{0}93.6 \pm \phantom{0}6.5}$ \\
HalfCheetah & $\phantom{0}54.0 \pm 29.6$ & $\phantom{00}1.7 \pm \phantom{0}0.1$ & $\phantom{0}28.9 \pm 31.6$ & $\phantom{0}49.2 \pm 35.2$ & $\mathbf{\phantom{0}87.6 \pm \phantom{0}2.6}$ \\
Hopper & $\phantom{0}60.7 \pm 19.6$ & $\mathbf{\phantom{0}96.0 \pm \phantom{0}3.4}$ & $\phantom{0}75.3 \pm 14.7$ & $\phantom{0}55.5 \pm 22.5$ & $\phantom{0}68.2 \pm 12.8$ \\
Walker2d & $\phantom{0}72.9 \pm 12.4$ & $\phantom{00}9.6 \pm \phantom{0}4.9$ & $\phantom{0}82.1 \pm \phantom{0}1.7$ & $\phantom{0}44.3 \pm 32.5$ & $\mathbf{\phantom{0}83.2 \pm \phantom{0}0.9}$ \\
InvertedPendulum & $\phantom{0}97.0 \pm \phantom{0}5.2$ & $\mathbf{100.0 \pm \phantom{0}0.0}$ & $100.0 \pm \phantom{0}0.0$ & $\phantom{0}89.4 \pm 26.0$ & $\phantom{0}99.3 \pm \phantom{0}1.7$ \\
InvertedDoublePendulum & $100.0 \pm \phantom{0}0.0$ & $\mathbf{100.1 \pm \phantom{0}0.0}$ & $100.0 \pm \phantom{0}0.0$ & $\phantom{0}95.5 \pm 10.7$ & $\phantom{0}98.4 \pm \phantom{0}4.0$ \\
Pusher & $\mathbf{\phantom{0}98.8 \pm \phantom{0}0.8}$ & $\phantom{0}97.9 \pm \phantom{0}1.2$ & $\phantom{0}98.0 \pm \phantom{0}1.1$ & $\phantom{0}97.9 \pm \phantom{0}1.5$ & $\phantom{0}98.2 \pm \phantom{0}2.2$ \\
Reacher & $102.2 \pm \phantom{0}1.0$ & $\mathbf{102.7 \pm \phantom{0}0.6}$ & $102.5 \pm \phantom{0}0.6$ & $102.2 \pm \phantom{0}1.2$ & $102.6 \pm \phantom{0}1.0$ \\
Swimmer & $\phantom{0}21.6 \pm \phantom{0}5.0$ & $\mathbf{\phantom{0}70.7 \pm 14.4}$ & $\phantom{0}16.0 \pm \phantom{0}2.3$ & $\phantom{0}22.4 \pm \phantom{0}4.1$ & $\phantom{0}20.3 \pm \phantom{0}2.4$ \\
\textbf{Medium Average} & $\phantom{0}74.5 \pm 27.1$ & $\phantom{0}62.4 \pm 44.1$ & $\phantom{0}76.7 \pm 30.4$ & $\phantom{0}64.7 \pm 32.0$ & $\mathbf{\phantom{0}81.7 \pm 25.1}$ \\
\midrule
\multicolumn{6}{c}{\textbf{Simple Dataset}} \\
\midrule
Humanoid & $\phantom{0}62.3 \pm \phantom{0}3.2$ & $\mathbf{\phantom{0}69.1 \pm \phantom{0}0.9}$ & $\phantom{0}65.3 \pm \phantom{0}3.4$ & $\phantom{0}64.0 \pm \phantom{0}3.8$ & $\phantom{0}62.1 \pm \phantom{0}4.9$ \\
Ant & $\phantom{0}84.4 \pm \phantom{0}5.2$ & $\phantom{0}14.8 \pm 30.9$ & $\phantom{0}88.0 \pm \phantom{0}4.5$ & $\phantom{0}84.2 \pm \phantom{0}6.1$ & $\mathbf{\phantom{0}97.7 \pm \phantom{0}0.0}$ \\
HalfCheetah & $\phantom{0}54.0 \pm \phantom{0}0.0$ & $\phantom{0}13.1 \pm 11.2$ & $\phantom{0}54.5 \pm \phantom{0}0.7$ & $\phantom{0}55.5 \pm \phantom{0}0.6$ & $\mathbf{\phantom{0}60.7 \pm \phantom{0}0.5}$ \\
Hopper & $\phantom{0}64.2 \pm 23.1$ & $\mathbf{\phantom{0}85.1 \pm \phantom{0}0.5}$ & $\phantom{0}73.9 \pm 16.4$ & $\phantom{0}62.6 \pm 22.0$ & $\phantom{0}65.1 \pm 17.9$ \\
Walker2d & $\phantom{0}54.0 \pm 15.7$ & $\phantom{00}4.4 \pm \phantom{0}0.4$ & $\mathbf{\phantom{0}64.6 \pm \phantom{0}3.4}$ & $\phantom{0}55.0 \pm 11.5$ & $\phantom{0}63.5 \pm \phantom{0}5.2$ \\
\textbf{Simple Average} & $\phantom{0}63.8 \pm 12.4$ & $\phantom{0}37.3 \pm 37.0$ & $\phantom{0}69.3 \pm 12.5$ & $\phantom{0}64.3 \pm 11.9$ & $\mathbf{\phantom{0}69.8 \pm 15.7}$ \\
\midrule
\textbf{Total Average} & $\phantom{0}71.7 \pm 24.3$ & $\phantom{0}53.3 \pm 39.7$ & $\phantom{0}71.3 \pm 26.4$ & $\phantom{0}66.9 \pm 25.2$ & $\mathbf{\phantom{0}77.1 \pm 24.2}$ \\
\bottomrule
\end{tabular}}
\end{table*}

\begin{figure}[t]
  \centering
  \begin{subfigure}[t]{0.49\textwidth}
    \centering
    \includegraphics[width=\linewidth]{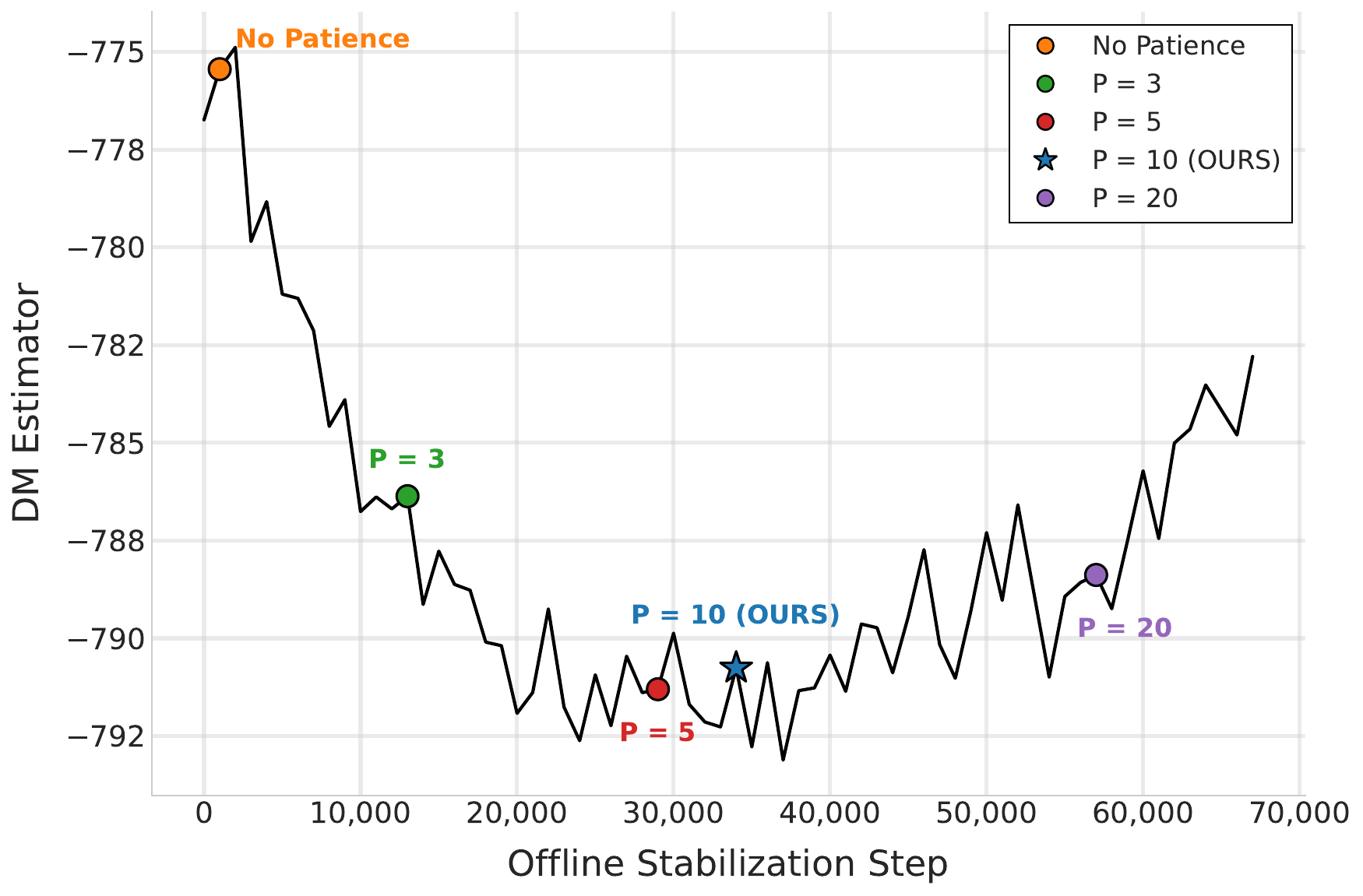}
    \caption{The DM estimator ($\widehat{J}_{\mathrm{DM}}$) during a stabilization phase. Markers indicate termination points chosen by our stopping criterion for different $P$ values.}

  \end{subfigure}\hfill
  \begin{subfigure}[t]{0.49\textwidth}
    \centering
    \includegraphics[width=\linewidth]{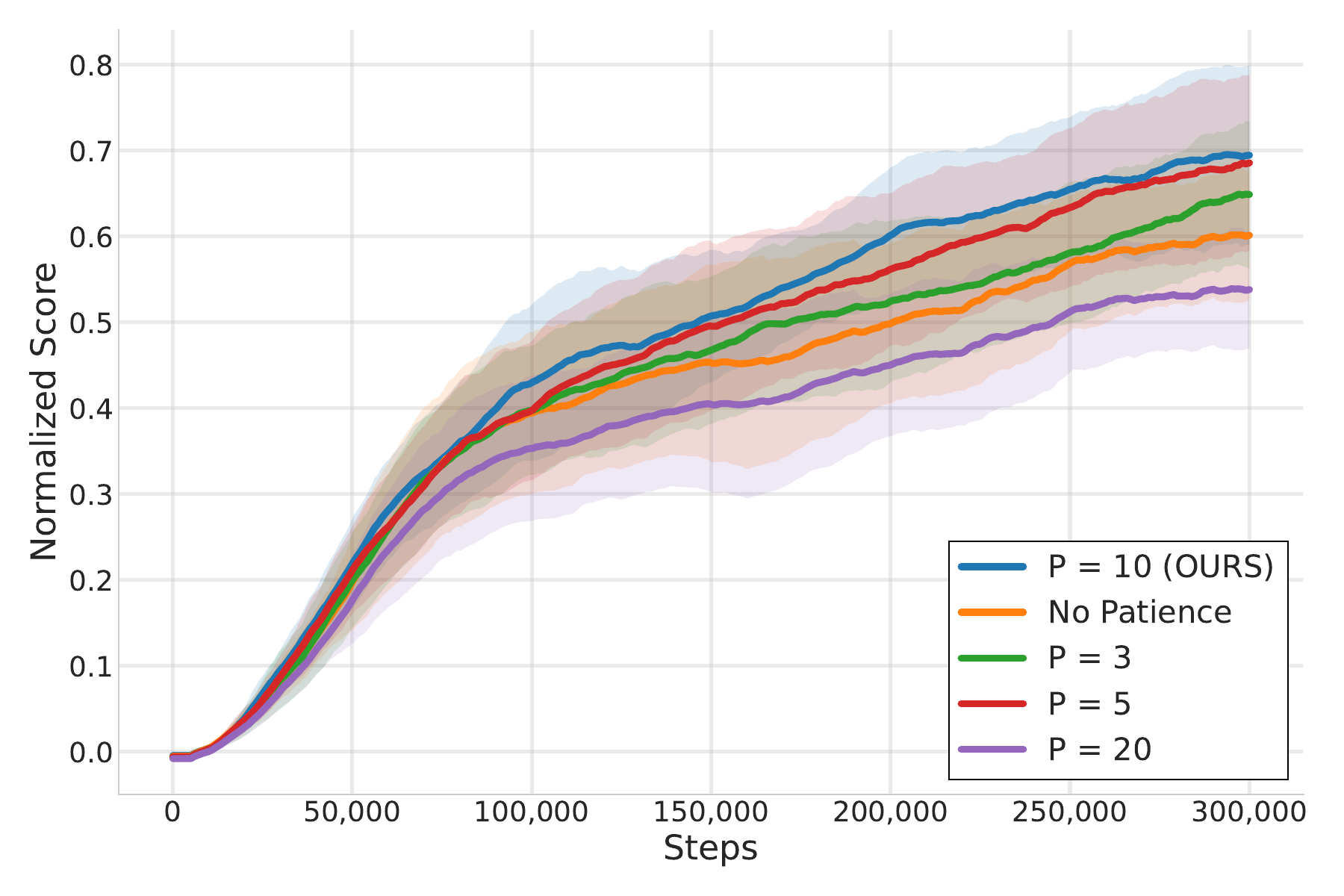}
    \caption{Aggregate end-to-end learning curves (normalized score) during training across different patience values.}

  \end{subfigure}

  \caption{\textbf{Ablation on the patience hyperparameter $P$.} \textbf{(a)} The evolution of the DM estimator during a single offline stabilization phase. Increasing $P$ prevents premature termination due to transient noise, but excessive patience ($P=20$) allows the estimator to diverge. \textbf{(b)} Aggregated performance shows the algorithm is robust to reasonable choices of $P \in \{3, 5, 10\}$, while extreme values (no patience or excessive patience) degrade end-to-end performance.}
  \label{fig:patience}
\end{figure}

\textbf{Q2: Is \nomealgo~robust to the choice of the patience hyperparameter?}
We analyze the sensitivity of \nomealgo~to the patience parameter $P$, which controls the termination and thus the length of each stabilization phase. In RL, the objective function, and specifically an OPE estimator, are notoriously noisy due to function approximation errors and stochastic action sampling. We hypothesize that $P$ acts as a crucial ``smoothing'' mechanism that allows the algorithm to distinguish between transient stochastic noise and true divergence of the DM estimator.

We first examine the intra-phase behavior of the actor-aligned objective $\widehat{J}_{\mathrm{DM}}$ across different patience values (Figure~\ref{fig:patience}a). Without patience ($P=1$), the stabilization phase is highly brittle; it terminates at the first noisy update, even when the underlying trend remains positive. As $P$ increases to 3, the process gains stability, allowing the critic to bypass local minima. We find that values of $P=5$ and $P=10$ offer an optimal balance, providing a sufficient look-ahead to confirm stagnation or divergence. Conversely, excessive patience ($P \geq 20$) proves counterproductive, as the stabilization phase continues long after the DM estimator has plateaued and started diverging, leading to computational waste and potential overfitting to the offline dataset.

To further evaluate the impact on end-to-end performance, we conducted full training runs across a range of patience values (Figure~\ref{fig:patience}b). The results confirm robust performance: reasonable values ($P \in \{3, 5, 10\}$) yield very similar, high-quality results. However, performance degrades at the extremes. Removing patience entirely collapses the update frequency toward the SACfD ($UTD \approx 1$), while high patience replicates the failure modes of fixed-length stabilization by forcing updates that no longer contribute to policy improvement. Comparing these findings with the results in Figure~\ref{fig:speq_brittle}, we argue that while some degree of patience is necessary to filter RL noise, \nomealgo~is more robust than SPEQ and does not require task-specific tuning of $P$.

\textbf{Q3: How does the algorithm distribute offline updates temporally across the training process?}
\label{sec:offline_epochs}
To investigate the relationship between adaptive stabilization and training dynamics, we analyze the OfflineUpdates metric, defined as the number of gradient updates performed during each stabilization phase as a function of environment steps.

\begin{figure}[t]
    \centering
        \includegraphics[width=0.5\textwidth]{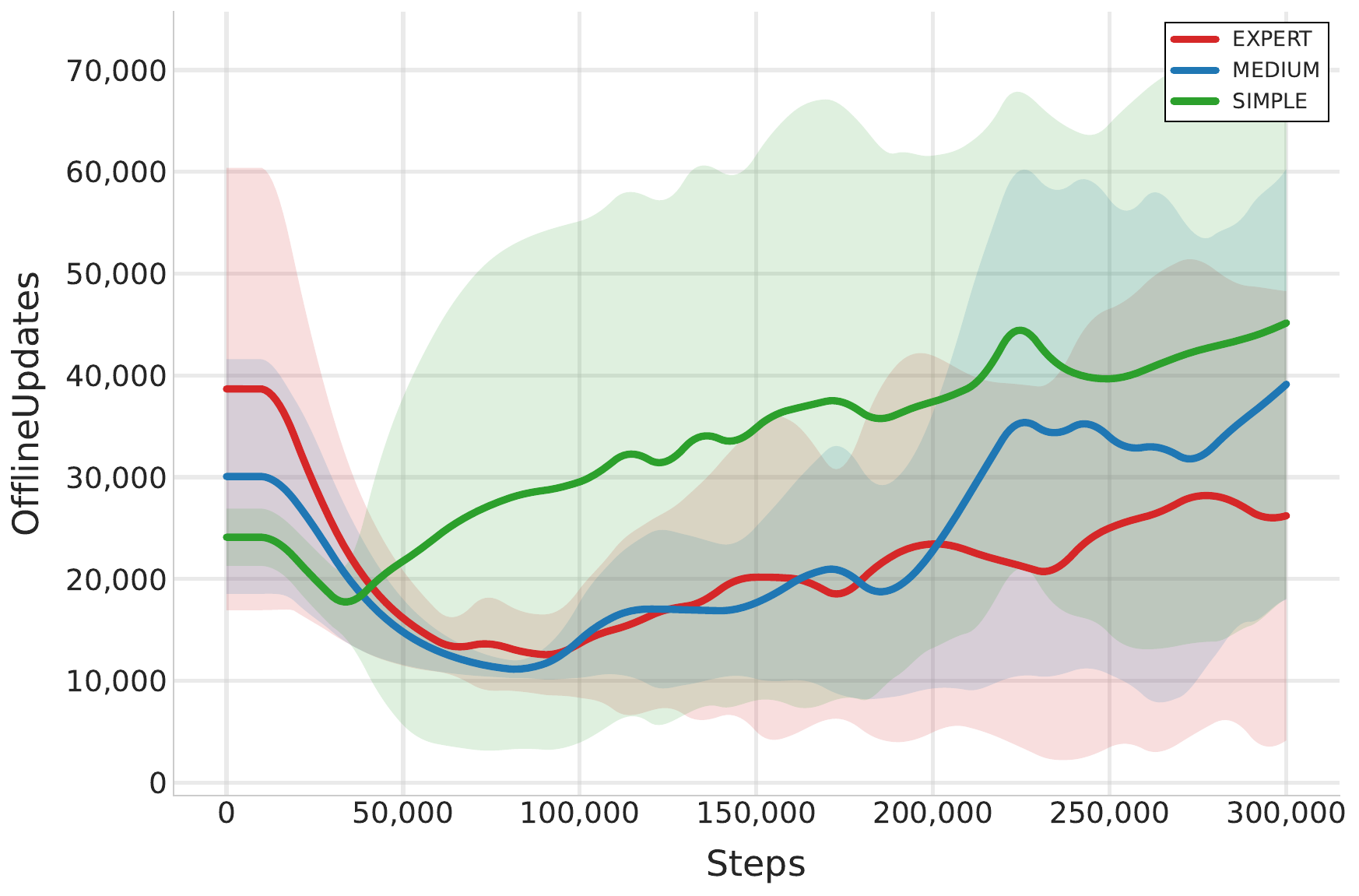}
    \caption{Distribution of offline updates (\textit{OfflineUpdates}) across training. The total average reveals a characteristic U-shaped profile: a high volume of updates in the early stages, a rapid decrease during mid-training, and a gradual resurgence in the late stages.}
    \label{fig:offline_epochs}
\end{figure}
Figure~\ref{fig:offline_epochs} shows the average distribution across all environments and datasets. The plot exhibits a distinct U-shaped profile characterized by three phases. In the \textbf{early stage} (0--30k steps), the algorithm allocates a high volume of offline updates (approximately 30k--32k offline updates per phase). This behavior aligns with the need for the critic to rapidly incorporate prior data to establish an initial value landscape; given that the policy is initially near-random, there is substantial capacity for critic improvement, and the OPE signal reflects the utility of additional updates. 
In the \textbf{middle stage} (30k--150k steps), the number of offline updates decreases to a minimum of about 21k. During this phase, as the policy begins to specialize, the critic quickly encounters diminishing returns with respect to the current samples in the buffer. At this point in training, overfitting may be a risk as the data distribution in the buffer increasingly mirrors the evolving policy. The OPE criterion identifies this saturation and terminates stabilization phases early. 
Finally, in the \textbf{late stage} (150k--300k steps), offline updates gradually increase to approximately 35k. As the policy approaches convergence, fine-grained value refinement becomes necessary; the critic requires additional updates that drive the final stages of policy improvement.
In the Supplementary Materials (Section~\ref{supp:experiments}), Figure~\ref{supp:offline_epochs} shows how \nomealgo~dynamically allocates the number of offline updates for each task.

\textbf{Q4: What is the computational efficiency of SOPE compared to baselines?}
\label{sec:compute} Table \ref{tab:summary} compares cumulative TFLOPs, wall-clock training time, and final normalized scores. While the normalized scores represent an average across 10 seeds for all 25 environment-dataset combinations, TFLOPs and wall-clock time are measured in an isolated setting to ensure a controlled evaluation. We selected the HalfCheetah Expert task for these measurements because its state-action space provides a representative medium complexity relative to other environments. In this setting, given that fluctuations in training time are typically negligible and due to the different initialization, all baseline models were evaluated using a single seed. In contrast, as SOPE is the only dynamic algorithm, its results have been averaged across 3 seeds to provide a more stable performance metric.

\nomealgo~achieves the highest normalized score ($77.94 \pm 25.11$) while requiring only 318.8 TFLOPs and 64 minutes of training time, on average, per task. This corresponds to a $36.8\times$, $8.2\times$, and $2.8\times$ reduction in TFLOPs relative to RLPD, Cal-QL, and SPEQ O2O, respectively, while simultaneously improving final performance by 45.6\%, 13.8\%, and 7.8\%, respectively. Compared to the SACfD baseline, \nomealgo~requires a $3.2\times$ increase in compute (479.8 vs.\ 152.3 TFLOPs) and $1.8\times$ increase in time, but with a 15.1\% performance improvement.

These results show that simply scaling up computation--such as through RLPD’s high UTD ratios or Cal-QL’s extensive offline pre-training--does not reliably yield proportional performance gains. By allocating resources only when the OPE signal indicates a clear benefit, \nomealgo~avoids the redundant computation inherent in fixed-schedule methods, maintaining an efficiency profile comparable to SACfD while significantly outperforming state-of-the-art baselines. Finally, in the Supplementary Materials (Section~\ref{sec:experiments}), Figure~\ref{supp:log_cost} compares the TFLOPs of each solution on a logarithmic scale, highlighting the orders-of-magnitude differences in computational cost.

\begin{table}[t]
\centering
\caption{Summary comparison across all methods. \textbf{TFLOPs} and \textbf{Time} are measured on the expert dataset. \textbf{Normalized Score} is the final score averaged across all 25 environment-dataset combinations and 10 seeds. \nomealgo~is the only algorithm that presents a variation in TFLOPs and Time due to its automatic allocation of offline updates during stabilization. It achieves the highest normalized score while consuming a fraction of the compute of RLPD and Cal-QL.}
\label{tab:summary}
\small % Slightly smaller to ensure fit
\resizebox{0.95\textwidth}{!}{%
\begin{tabular}{l *{4}{c} r@{\,$\pm$\,}l}
\toprule
 & \textbf{RLPD} & \textbf{Cal-QL} & \textbf{SPEQ O2O} & \textbf{SACfD} & \multicolumn{2}{c}{\textbf{\nomealgo~(OURS)}} \\
\midrule
TFLOPs & 11708.1 & 2618.2 & 897.0 & 152.3 & \textbf{318.8} & \textbf{18.7} \\
Time (min) & 1006 & 681 & 124 & 30 & \textbf{64} & \textbf{19.1} \\
Norm. Score & 53.54 $\pm$ 39.65 & 68.49 $\pm$ 25.69 & 72.29 $\pm$ 25.20 & 67.62 $\pm$ 25.56 & \textbf{77.94} & \textbf{25.11} \\
\bottomrule
\end{tabular}
}
\end{table}

\section{Conclusions}
\label{sec:conclusions}
In this work we introduced \nomealgo, a novel algorithm for online Reinforcement Learning with prior data that automates the computational schedule of offline stabilization phases by utilizing an actor-aligned Off-policy Policy Evaluation signal. By framing stabilization as an early-stopping problem, SOPE eliminates the need for manual, task-specific hyperparameter tuning and achieves state-of-the-art performance with a significantly reduced computational footprint, yielding up to a 36.8× reduction in TFLOPs compared to high-UTD baselines across 25 continuous control tasks. 

While our results demonstrate high robustness across the Minari benchmark, a potential limitation lies in the dependence on the validation data distribution; the reliability of the OPE signal is intrinsically tied to how well the held-out split represents the state-action space the policy will explore. Finally, in this paper we demonstrated that \nomealgo~is effective in the online RL with prior data setting. However, it remains unclear whether the same dynamic stabilization phase approach would generalize to an online-only setting, where no offline data is available.

% \label{sec:ack}

%%%%%%%%%%%%%%%%%%%%%%%%%%%%%%%%%%%%%%%%%%%%%%%%%%%%%%%%%%%%%%%%
%% NOTE: THIS MARKS THE END OF THE "MAIN TEXT"
%%%%%%%%%%%%%%%%%%%%%%%%%%%%%%%%%%%%%%%%%%%%%%%%%%%%%%%%%%%%%%%%

%%%%%%%%%%%%%%%%%%%%%%%%%%%%%%%%%%%%%%%%%%%%%%%%%%%%%%%%%%%%%%%%
%% Bibliography
%%%%%%%%%%%%%%%%%%%%%%%%%%%%%%%%%%%%%%%%%%%%%%%%%%%%%%%%%%%%%%%%
\bibliography{main}
\bibliographystyle{rlj}

%%%%%%%%%%%%%%%%%%%%%%%%%%%%%%%%%%%%%%%%%%%%%%%%%%%%%%%%%%%%%%%%
% AUTHOR: If your paper has no supplementary materials, you may 
%         comment out the line below, which creates the title for
%         the supplementary materials.
%%%%%%%%%%%%%%%%%%%%%%%%%%%%%%%%%%%%%%%%%%%%%%%%%%%%%%%%%%%%%%%%
\beginSupplementaryMaterials

\section{Pseudo-code of the \nomealgo~algorithm}

Algorithm~\ref{supp:sope-algo} provides pseudo-code for the SOPE algorithm.

\label{supp:algo}
\begin{algorithm}[H]
\caption{\nomealgo}
\label{supp:sope-algo}
\begin{algorithmic}[1]
\small
\STATE \textbf{Input:} Initial parameters $\theta, \phi$, Patience $P$, Eval Interval $E$
\WHILE{training}
    \STATE // \textbf{Online Phase}
    \STATE Interact with environment for $N$ steps and store in $\mathcal{D}$.
    \STATE Perform $N$ standard SAC updates with simmetric sampling on $\theta, \phi$.
    \STATE // \textbf{Adaptive Stabilization Phase}
    \STATE Partition $\mathcal{D} \to \mathcal{D}_{\text{train}}, \mathcal{D}_{\text{val}}$.
    \STATE $M \gets -\infty$, $p \gets 0$
    \WHILE{$p < P$}
        \STATE Perform $E$ updates on $Q_\phi$ using symmetric sampling from $\mathcal{D}_{\text{train}}$.
        \STATE $\widehat{J}_{DM} \gets \frac{1}{|\mathcal{D}_{\text{val}}|} \sum_{s \in \mathcal{D}_{\text{val}}} Q_\phi(s, a), a \sim \pi_\theta(\cdot|s)$
        \IF{$\widehat{J}_{DM} > M$}
            \STATE $M \gets \widehat{J}_{DM}$, $p \gets 0$
        \ELSE
            \STATE $p \gets p + 1$
        \ENDIF
    \ENDWHILE
\ENDWHILE
\end{algorithmic}
\end{algorithm}

\section{Extended Performance Comparison}
\label{supp:experiments}
\begin{figure}[htbp]
    \centering
    % --- First Subfigure ---
    \begin{subfigure}[b]{0.3\textwidth}
        \centering
        % Replace 'example-image-a' with your filename
        \includegraphics[width=\textwidth]{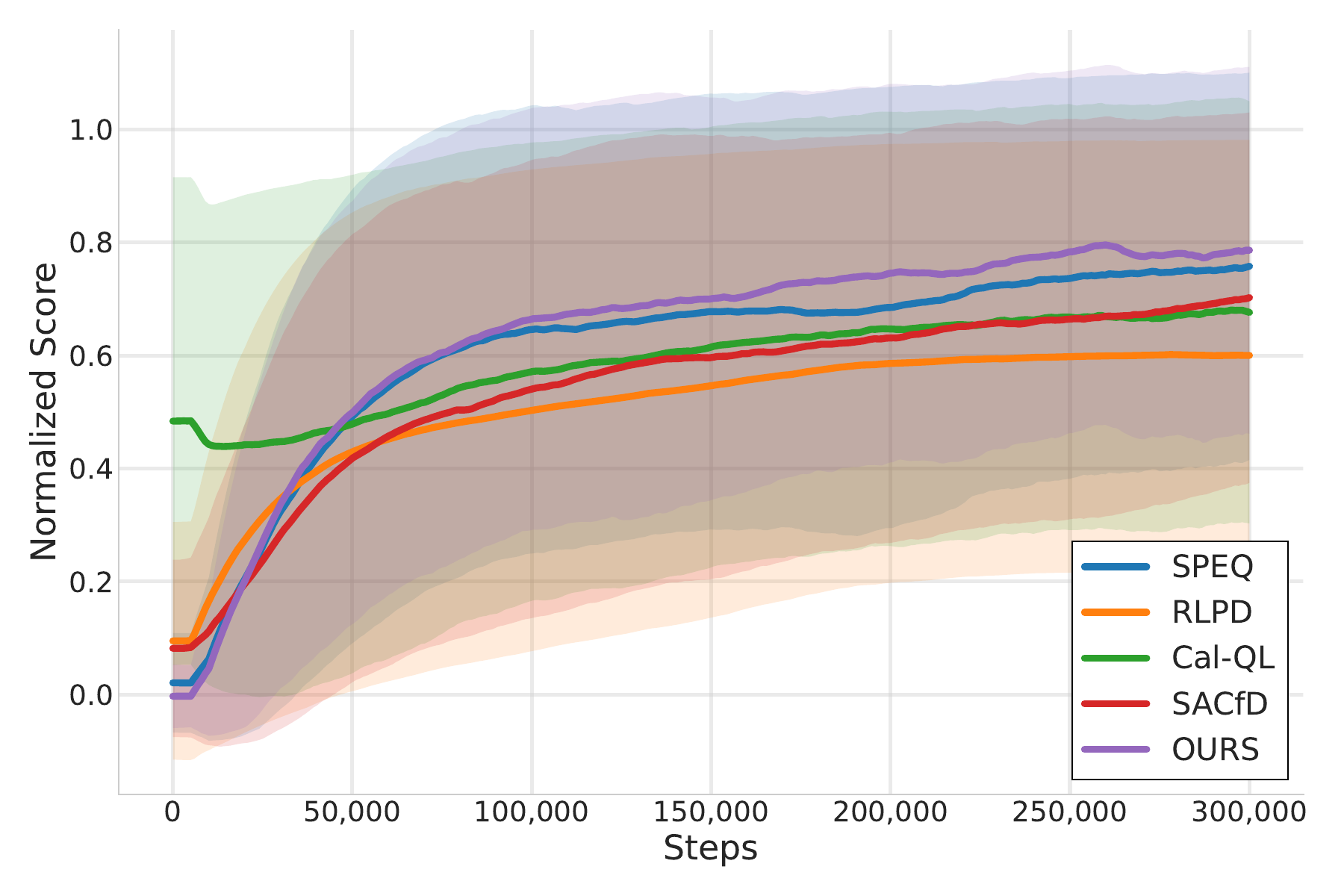}
        \caption{Expert}
        \label{fig:fig1}
    \end{subfigure}
    \hfill % Adds flexible space between images
    % --- Second Subfigure ---
    \begin{subfigure}[b]{0.3\textwidth}
        \centering
        % Replace 'example-image-b' with your filename
        \includegraphics[width=\textwidth]{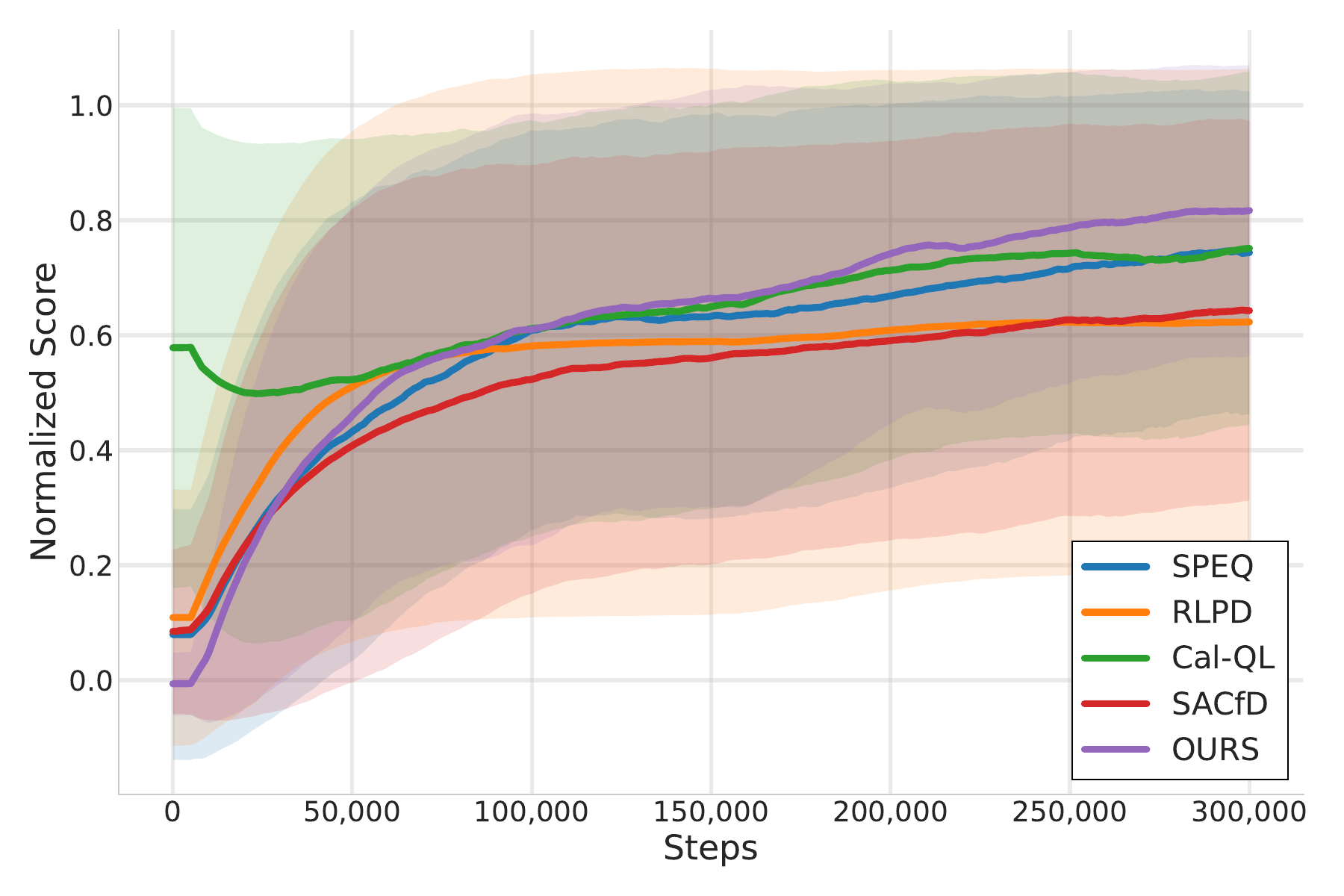}
        \caption{Medium}
        \label{fig:fig2}
    \end{subfigure}
    \hfill % Adds flexible space between images
    % --- Third Subfigure ---
    \begin{subfigure}[b]{0.3\textwidth}
        \centering
        % Replace 'example-image-c' with your filename
        \includegraphics[width=\textwidth]{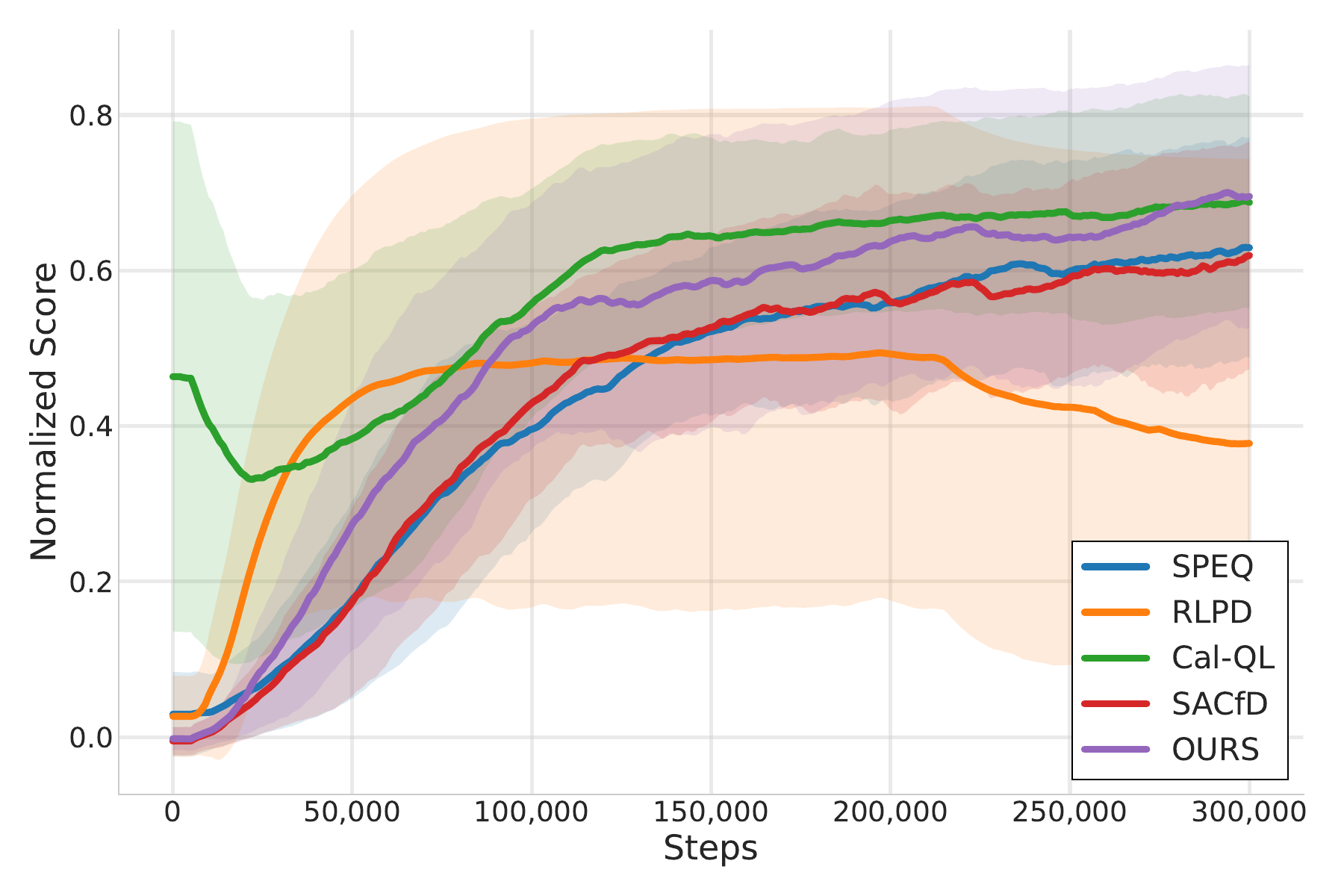}
        \caption{Simple}
        \label{fig:fig3}
    \end{subfigure}
    
    \caption{\textbf{Aggregated normalized scores across dataset qualities.} The plots show the aggregate end-to-end performance of our method against baselines on the (a) Expert, (b) Medium, and (c) Simple dataset splits. Shaded regions denote the standard deviation across 10 random seeds.}
    \label{supp:aggregated}
\end{figure}

\textbf{Q1: Is dynamic allocation of offline updates superior to fixed schedules?} In support of the results presented in Table~\ref{tab:results_all}, Figure~\ref{supp:aggregated} provides the aggregated results describing the results obtained by each algorithm for each task.

\begin{figure}[htbp]
    \centering
    
    % --- Expert Split ---
    \begin{subfigure}[b]{\textwidth}
        \centering
        \includegraphics[width=0.99\textwidth]{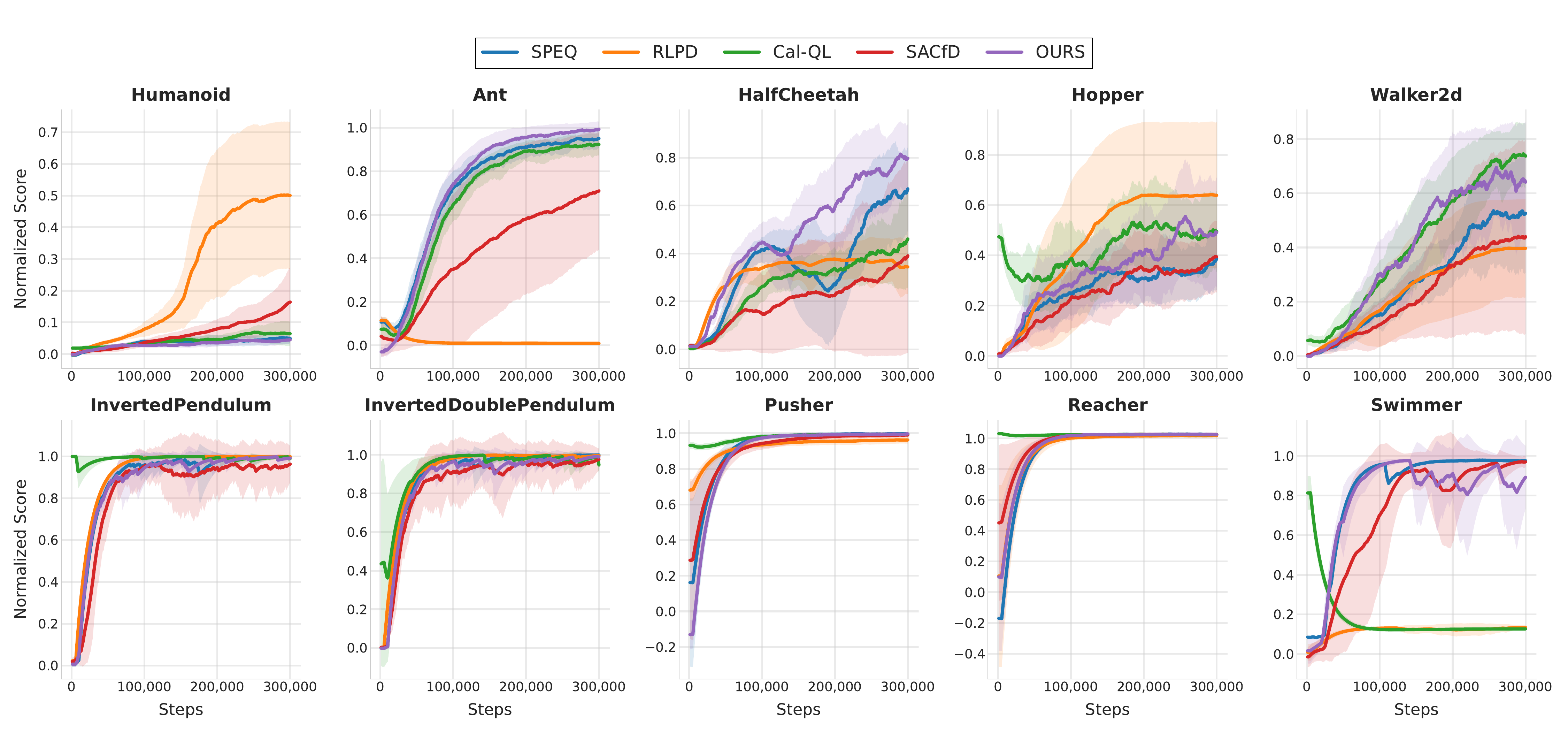}
        \caption{Expert}
        \label{fig:split_expert}
    \end{subfigure}
    
    \vspace{1em} % Adds vertical space between the stacked images
    
    % --- Medium Split ---
    \begin{subfigure}[b]{\textwidth}
        \centering
        \includegraphics[width=0.99\textwidth]{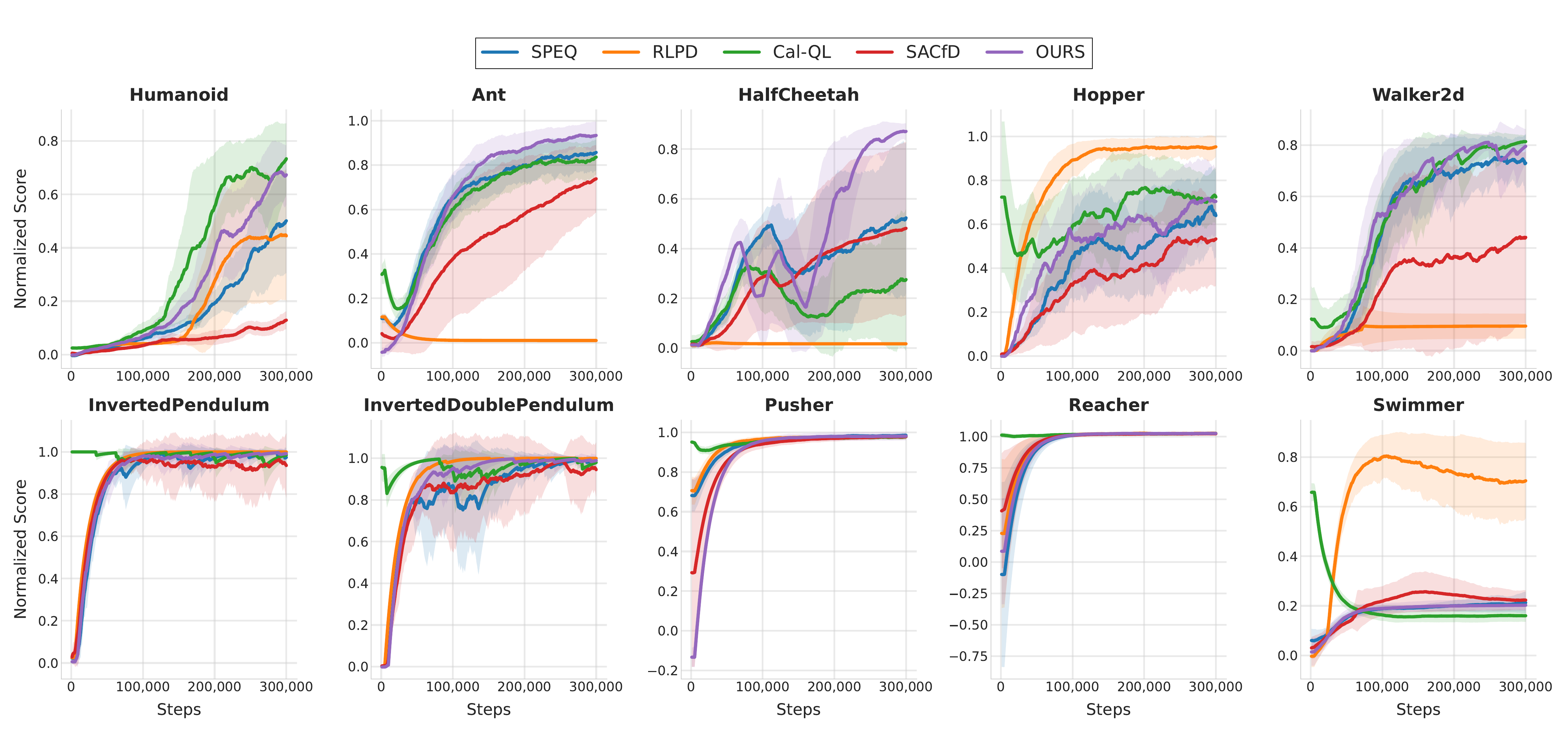}
        \caption{Medium}
        \label{fig:split_medium}
    \end{subfigure}
    
    \vspace{1em} % Adds vertical space between the stacked images
    
    % --- Simple Split ---
    \begin{subfigure}[b]{\textwidth}
        \centering
        \includegraphics[width=0.99\textwidth]{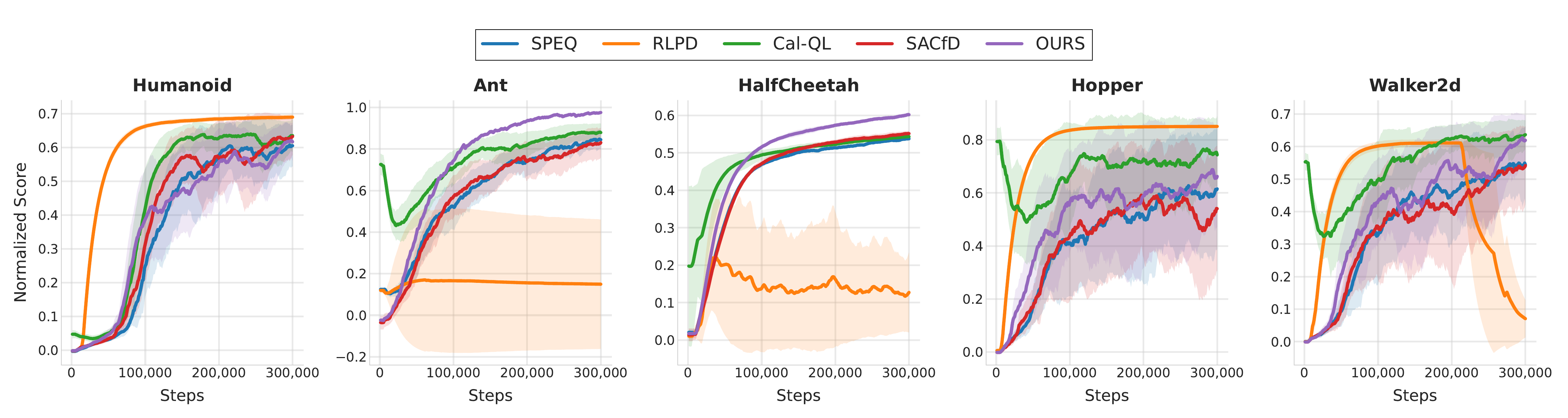}
        \caption{Simple}
        \label{supp:split_simple}
    \end{subfigure}

    \caption{\textbf{Individual learning curves per task and dataset quality.} Exploded views detailing the single learning trend for each algorithm and specific environment combination, across 10 random seeds. The results are grouped vertically by dataset quality: (a) Expert, (b) Medium, and (c) Simple. Our adaptive re-distribution method (OURS) shows robust or superior performance across most individual environments.}
    \label{supp:exploded_views}
\end{figure}

Furthermore, in Figure~\ref{supp:exploded_views}, we provide the exploded plots showing the individual learning trends for each algorithm and combination (environment, dataset).

% \textbf{Is \nomealgo~robust to the choice of the patience hyperparameter?}

\begin{figure}[htbp]
    \centering
    % --- First Subfigure ---
    \begin{subfigure}[b]{0.3\textwidth}
        \centering
        % Replace 'example-image-a' with your filename
        \includegraphics[width=\textwidth]{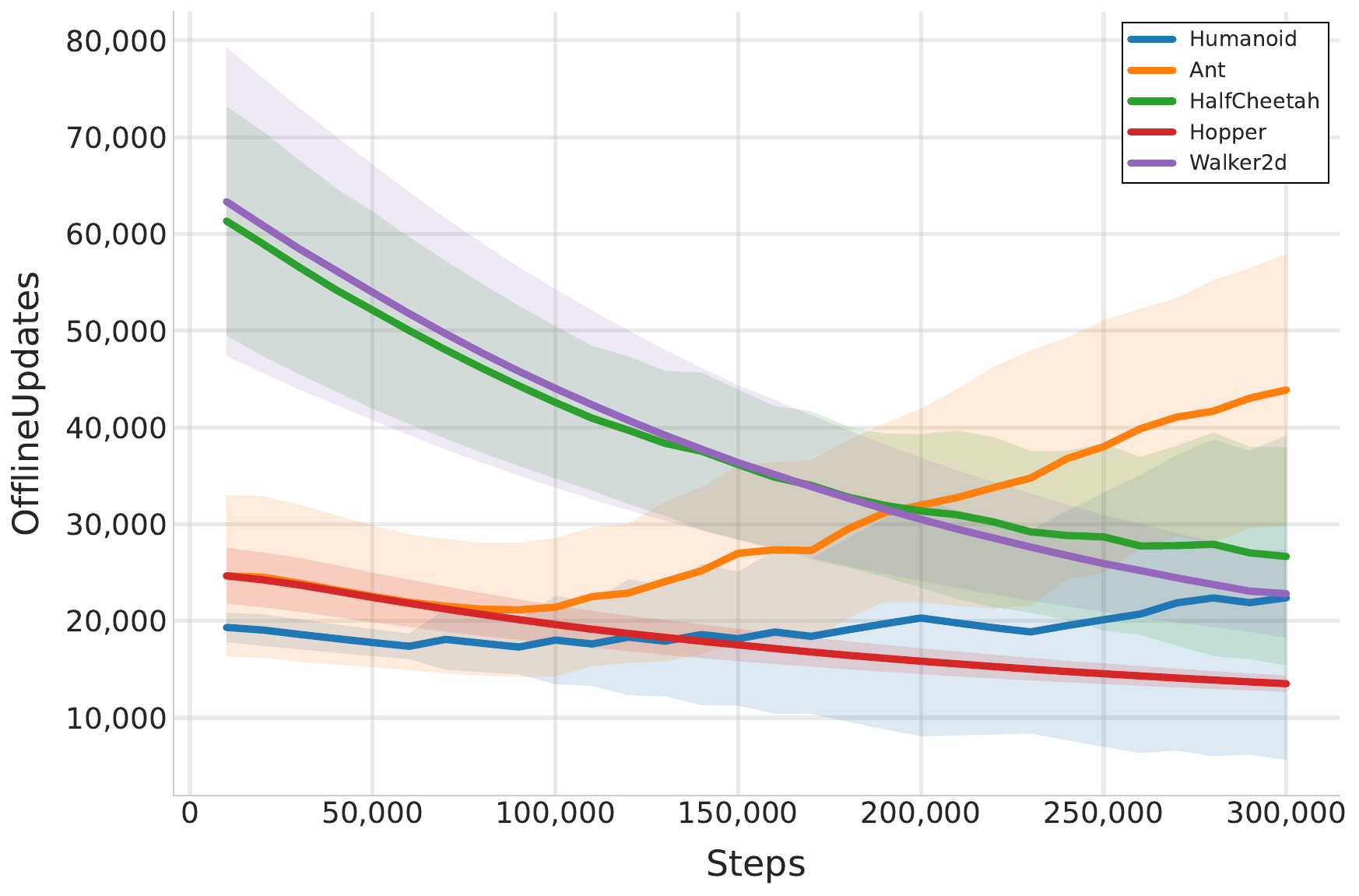}
        \caption{Expert}
    \end{subfigure}
    \hfill % Adds flexible space between images
    % --- Second Subfigure ---
    \begin{subfigure}[b]{0.3\textwidth}
        \centering
        % Replace 'example-image-b' with your filename
        \includegraphics[width=\textwidth]{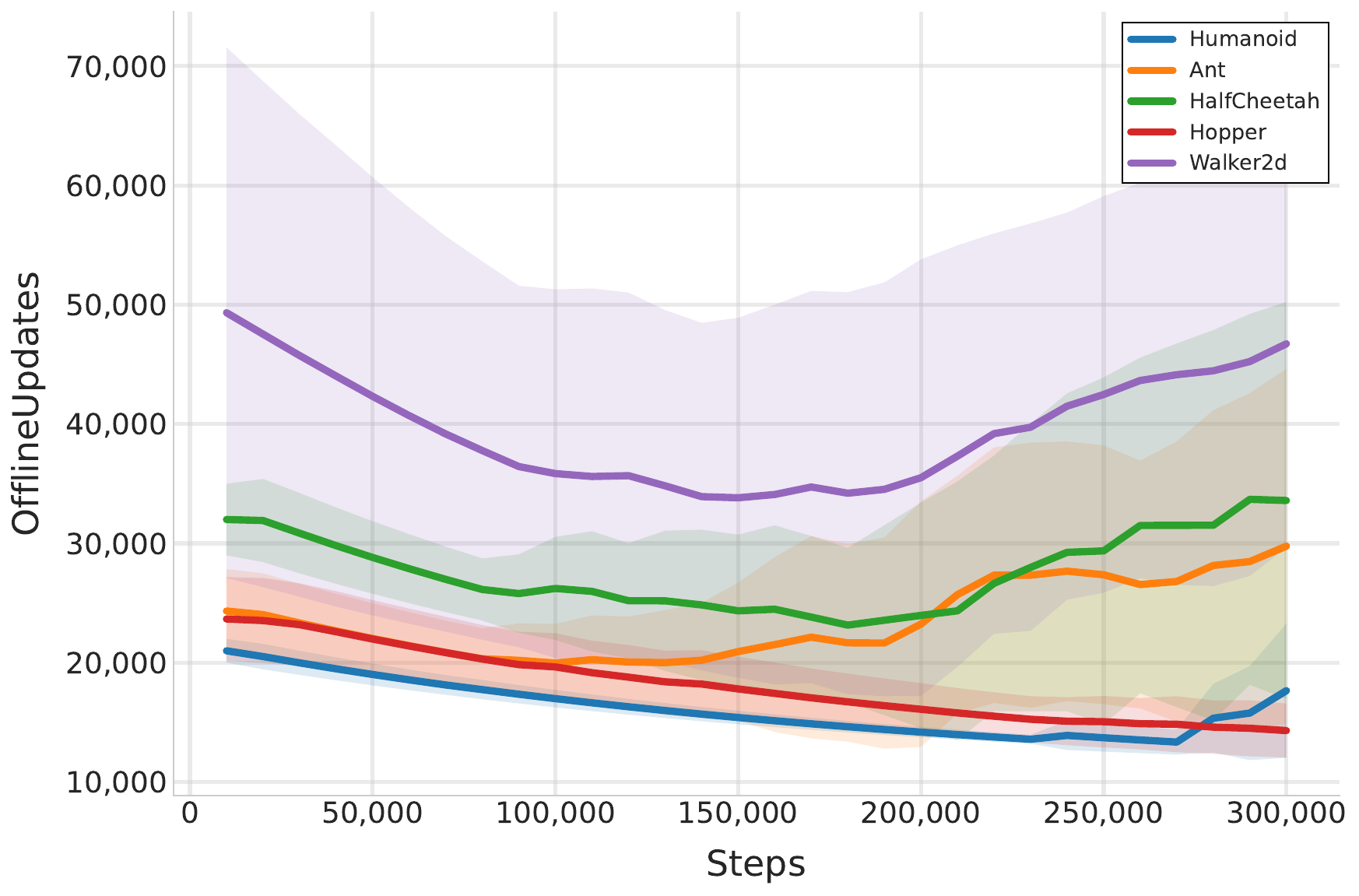}
        \caption{Medium}
        
    \end{subfigure}
    \hfill % Adds flexible space between images
    % --- Third Subfigure ---
    \begin{subfigure}[b]{0.3\textwidth}
        \centering
        % Replace 'example-image-c' with your filename
        \includegraphics[width=\textwidth]{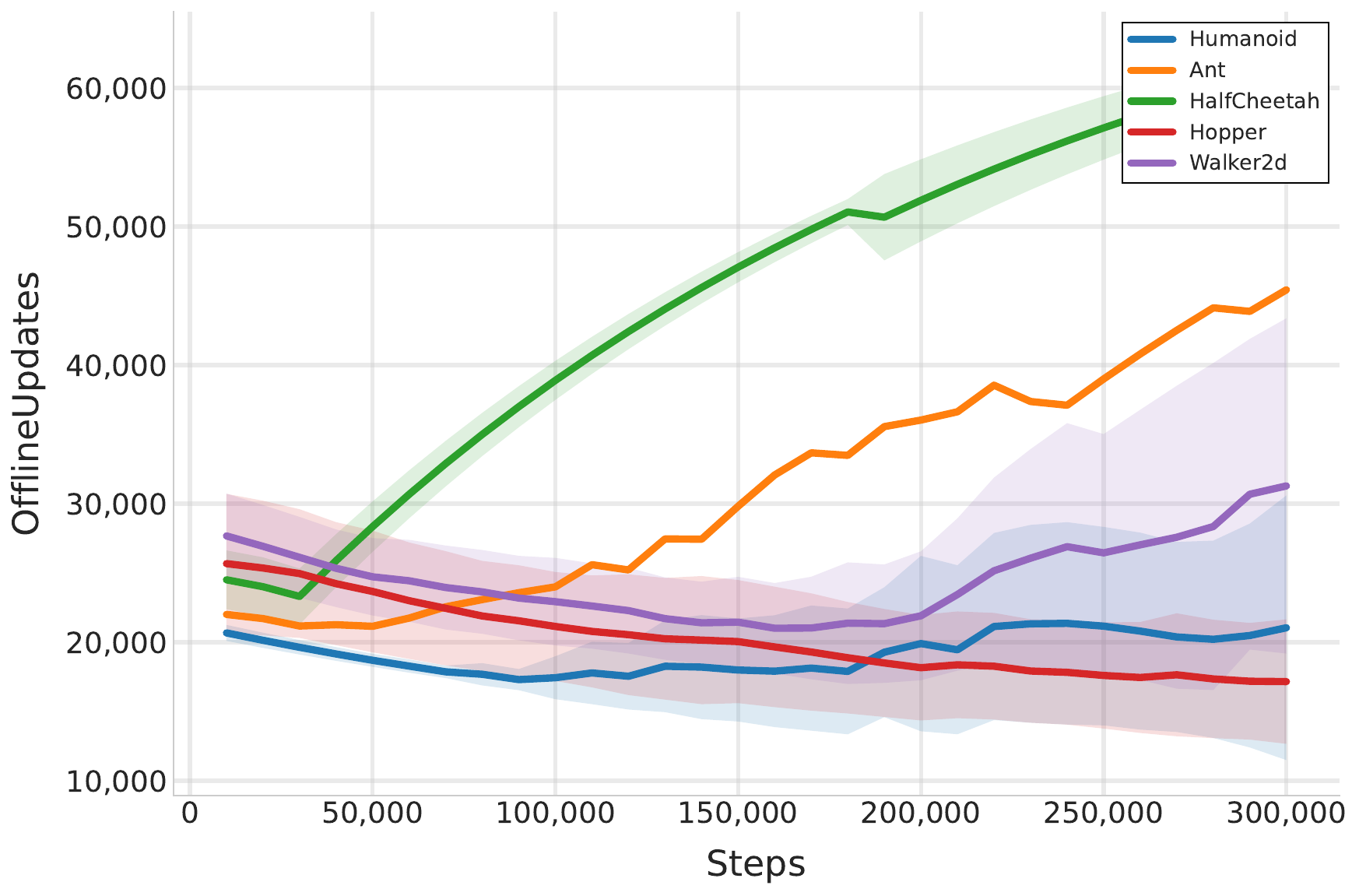}
        \caption{Simple}
    \end{subfigure}
    
    \caption{\textbf{Temporal distribution of offline updates per environment.} Evolution of the number of offline epochs during fine-tuning for the (a) Expert, (b) Medium, and (c) Simple datasets. The plots demonstrate how \nomealgo~dynamically adapts the length of offline stabilization phases over time, with the required number of updates varying significantly depending on the specific task and dataset quality.}
    \label{supp:offline_epochs}
\end{figure}

\textbf{Q3: How does the algorithm distribute offline updates temporally across the training process?} In support of the results shown in Figure~\ref{fig:offline_epochs}, in Figure~\ref{supp:offline_epochs}, we provide plots, broken down by tasks, showing how \nomealgo~dynamically chooses the duration of the offline stabilization phases in relation to the current environment.

\begin{figure}[t]
    \centering
    \includegraphics[width=0.65\textwidth]{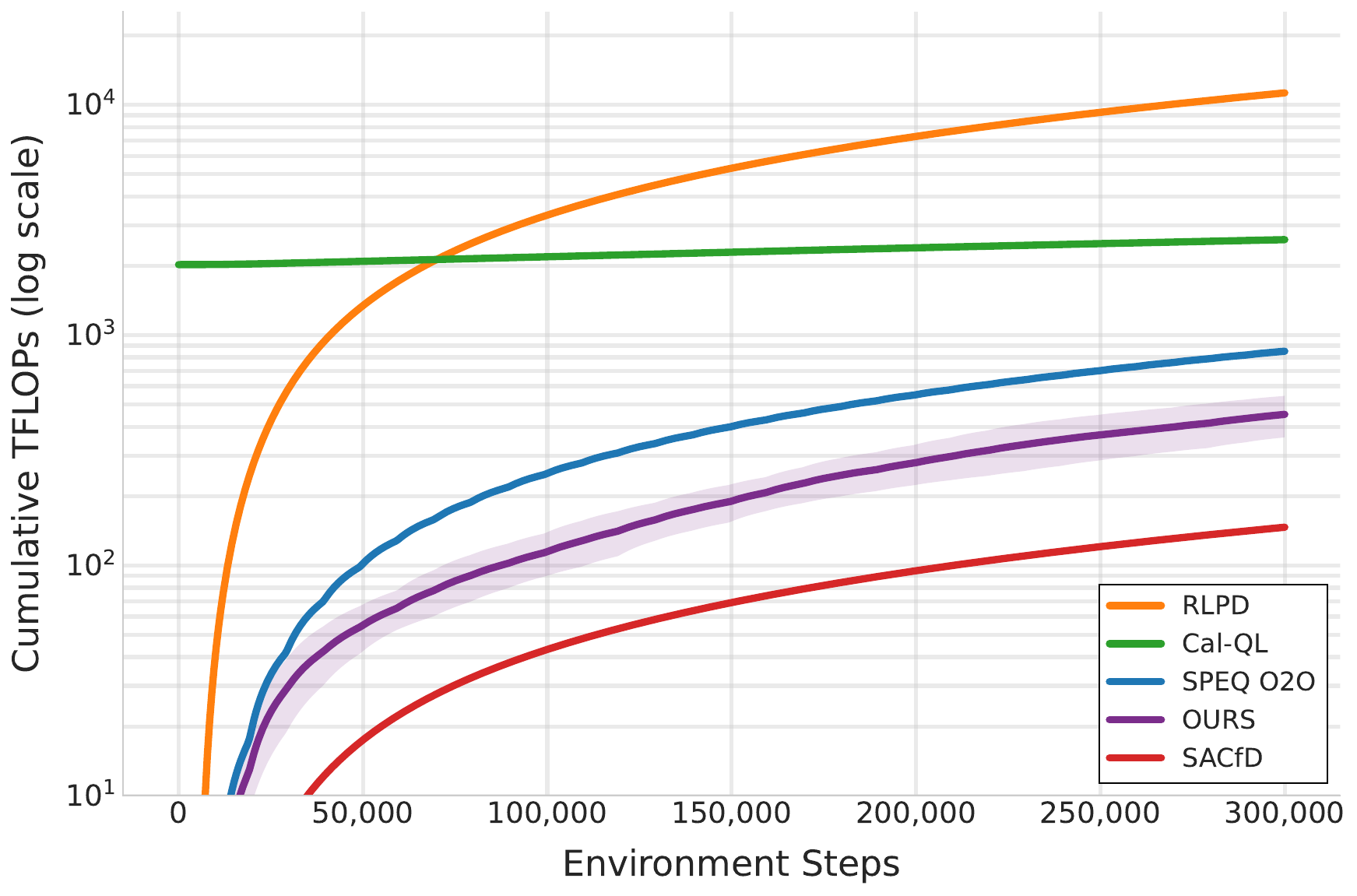}
    \caption{\textbf{Computational cost efficiency.} Cumulative TFLOPs (log scale) across environment steps. The plot demonstrates that our adaptive update distribution method (OURS) maintains a significantly lower computational footprint compared to high-update methods (RLPD) and extensive offline pre-training baselines (Cal-QL), remaining highly efficient throughout the fine-tuning process.}
    \label{supp:log_cost}
\end{figure}

\textbf{Q4: What is the computational efficiency of SOPE compared to baselines?} To support the findings in Table~\ref{tab:summary}, Figure~\ref{supp:log_cost} compares the computational cost of each algorithm in TFLOPs. We use a logarithmic scale to clearly illustrate the vast differences in magnitude.

\end{document}